\documentclass[letterpaper]{article} 
\usepackage{aaai2026}  
\usepackage{times}  
\usepackage{helvet}  
\usepackage{courier}  
\usepackage[hyphens]{url}  
\usepackage{graphicx} 
\usepackage{natbib}  
\usepackage{caption} 
\usepackage{booktabs}
\usepackage{amsfonts}
\usepackage{amsmath}
\usepackage{graphicx}
\usepackage{subcaption}
\usepackage{amssymb}
\usepackage{float}

\usepackage{newfloat}
\usepackage{listings}
\DeclareCaptionStyle{ruled}{labelfont=normalfont,labelsep=colon,strut=off} 
\floatstyle{ruled}
\newfloat{listing}{tb}{lst}{}
\floatname{listing}{Listing}
\title{CoMAD: A Multiple-Teacher Self-Supervised Distillation Framework}

\author{%
  Sriram Mandalika\textsuperscript{1\equalcontrib}\thanks{Corresponding Author.}, Lalitha V\textsuperscript{2\equalcontrib}\footnotemark[1]%
}

\affiliations{
    \textsuperscript{\rm 1}Department of Computational Intelligence,\\
    \textsuperscript{\rm 2}Department of Electronics and Communication Engineering\\
    Faculty of Engineering and Technology, SRM Institute of Science and Technology \\ 
    Kattankulathur, Tamil Nadu, 603203, India

    {mc9991, lv2876}@srmist.edu.in
}

\usepackage{bibentry}

\begin{document}

\maketitle

\begin{abstract}
Numerous self-supervised learning paradigms, such as contrastive learning and masked image modeling, learn powerful representations from unlabeled data but are typically pretrained in isolation, overlooking complementary insights and yielding large models that are impractical for resource-constrained deployment. To overcome these challenges, we introduce Consensus-oriented Masked Distillation - \textit{CoMAD}, a lightweight, parameter-free framework that unifies knowledge from multiple current state-of-the-art self-supervised Vision Transformers into a compact student network. CoMAD distills from three pretrained ViT-Base teachers, MAE, MoCo v3, and iBOT, each offering distinct semantic and contextual priors. Rather than naively averaging teacher outputs, we apply asymmetric masking, the student sees only 25 \% of patches while each teacher receives a progressively lighter, unique mask, forcing the student to interpolate missing features under richer contexts. Teacher embeddings are aligned to the student’s space via a linear adapter and layer normalization, then fused through our joint consensus gating, which weights each token by combining cosine affinity with inter-teacher agreement. The student is trained with dual-level KL divergence on visible tokens and reconstructed feature maps, capturing both local and global structure. On ImageNet-1K, CoMAD’s ViT-Tiny achieves 75.4 \% Top-1, an increment of 0.4 \% over the previous state-of-the-art. In dense-prediction transfers, it attains 47.3 \% mIoU on ADE20K, and 44.5 \% box average precision and 40.5 \% mask average precision on MS-COCO, establishing a new state-of-the-art in compact SSL distillation.
\end{abstract}

\section{Introduction}

Self‐supervised learning (SSL) \cite{9086055, Zbontar2021BarlowTS, Liu2020SelfSupervisedLG} has emerged as a foundation for modern computer vision. By training an encoder on a pretext task using vast unlabeled data, SSL yields versatile, task‐agnostic features without costly manual annotations, which can then be fine‐tuned on diverse downstream tasks. Two dominant paradigms have taken hold: contrastive learning (CL) \cite{He2019MomentumCF, Chen2020ImprovedBW, Zhang2023PatchLevelCW}, which maximizes agreement between augmented views of the same image; and masked image modeling (MIM) \cite{Chen2020ASF, He2021MaskedAA, Chen2022ContextAF}, which reconstructs corrupted patches by leveraging global context. Building on Masked Autoencoders (MAE) \cite{He2021MaskedAA}, recent works such as i‐MAE \cite{Zhang2022iMAEAL}, ADMA \cite{Liu2023ContinualMAEAD}, MoCE \cite{Liu2024TaskcustomizedMA}, and SCE‐MAE \cite{Yin2024SCEMAESC} have pushed the boundaries of MIM, yet they all depend on large, resource‐intensive backbones.

\begin{figure}[t]
  \centering
  \includegraphics[width=\columnwidth]{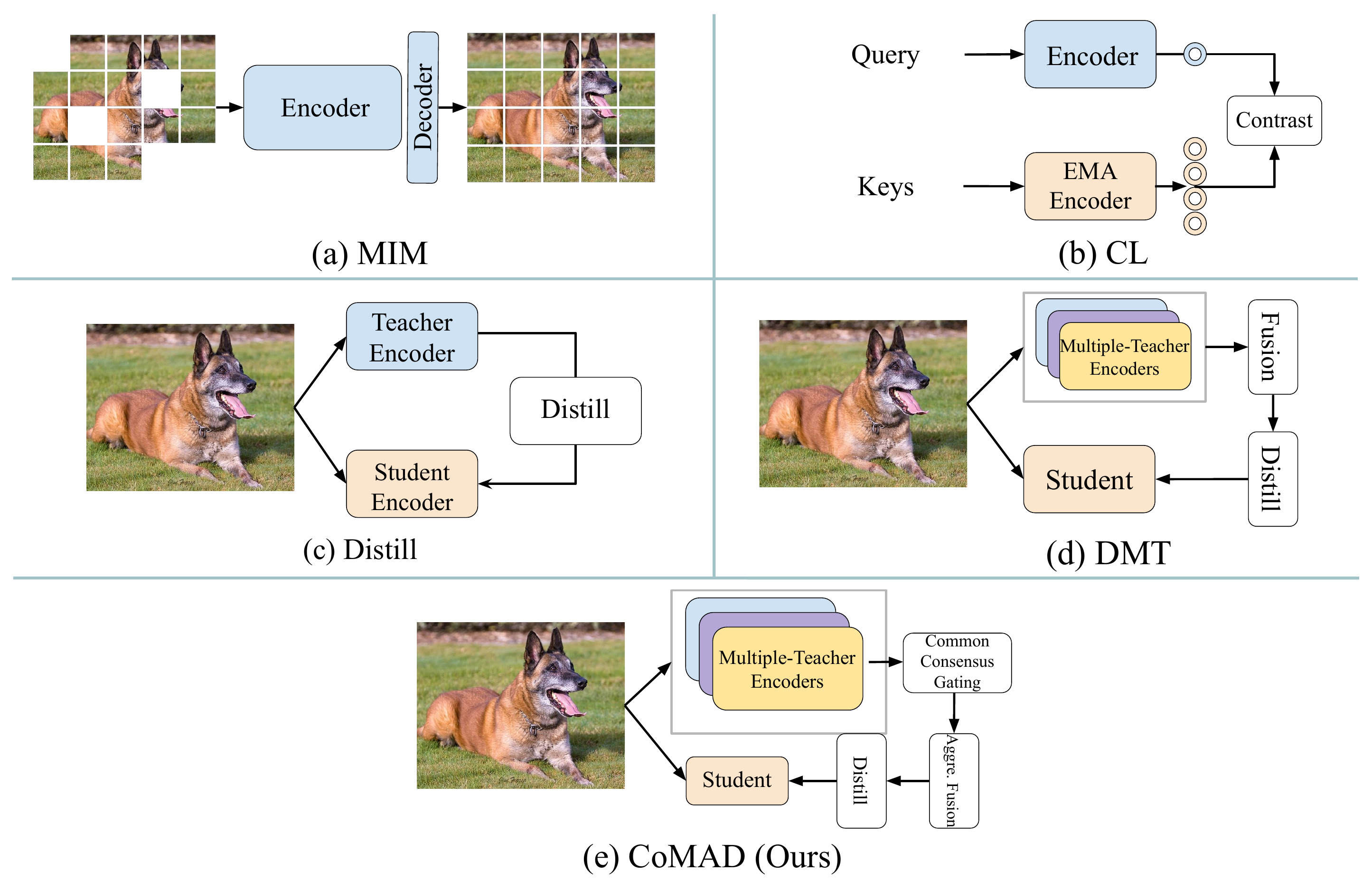}
  \caption{Various paradigms of self-supervised learning, in-
cluding (a) Masked Image Modeling (MIM), (b) Contrastive
Learning (CL), (c) Distillation, (d) DMT and (e) Our CoMAD.}
  \label{fig:comad_single}
\end{figure}

Knowledge distillation (KD) \cite{Hinton2015DistillingTK, Mansourian2025ACS} offers an elegant remedy by transferring knowledge from a large teacher to a lightweight student, achieving both compression and performance retention. KD has been successfully applied in supervised image classification \cite{Hinton2015DistillingTK}, object detection \cite{He2017MaskR, Chen2017LearningEfficientOBJKD}, semantic segmentation \cite{Li2017KDforSeg, Park2019RelationalKD}, and even large‐scale language models \cite{Sanh2019DistilBERTAD}. However, most KD methods focus on supervised tasks or single‐teacher SSL distillation \cite{Touvron2020TrainingDI, Ren2023TinyMIMAE}, leaving the rich, complementary insights from multiple SSL teachers unexplored. A few supervised multi‐teacher works \cite{Anil2018LargeSD, Daquan2020RethinkingBS} have attempted to balance heterogeneous teacher signals, but they rely on fixed or parametric weighting schemes that may conflict.

Despite these advances, distilling from multiple SSL teachers poses unique hurdles. Each teacher, pretrained under a different self‐supervised objective, encodes complementary yet potentially conflicting representations, and naive averaging can confuse the student. Supervised multi‐teacher KD methods \cite{Anil2018LargeSD, Daquan2020RethinkingBS} rely on fixed or parametric weighting schemes that are often brittle or parameter‐heavy, while single‐teacher SSL distillation approaches \cite{Touvron2020TrainingDI, Ren2023TinyMIMAE} sidestep fusion altogether and miss out on ensemble benefits. Consequently, a lightweight, parameter‐free fusion mechanism that dynamically reconciles divergent teacher signals at the token level remains an open challenge.

To address these challenges, we present \underline{Co}nsensus-oriented \underline{MA}sked \underline{D}istillation (CoMAD), a lightweight, parameter-free gating framework that distills knowledge from multiple ViT-Base SSL teachers into a compact ViT-Tiny student. CoMAD employs asymmetric masking to generate diverse partial views, aligns teacher embeddings to the student space via lightweight adapters, and uses a joint consensus gating mechanism to dynamically fuse teacher signals at each token. We then train the student with dual-level KL divergence losses on both visible tokens and reconstructed feature maps, ensuring it captures both local semantics and global structure. Our main contributions are:
\begin{enumerate}
  \item We propose a \textbf{multi-teacher distillation schema} that harnesses three ViT-Base models pretrained via MAE \cite{He2021MaskedAA}, MoCo v3 \cite{Chen2021AnES}, and iBOT \cite{Zhou2021iBOTIB}, each under a distinct self-supervised paradigm, to capture truly complementary representations.
  \item We designed a \textbf{parameter-free joint consensus gating mechanism} that adaptively weights each teacher per token based on student–teacher affinity and inter-teacher agreement, enabling conflict-aware fusion without introducing extra learnable parameters.
  \item We develop an \textbf{asymmetric masked distillation protocol} in which the student is heavily masked while each teacher receives a lighter, distinct mask, ensuring teachers observe more context than the student and fostering robust knowledge transfer.
  \item We present a \textbf{dual-level soft fusion strategy} that combines weighted teacher embeddings at both token and spatial feature-map levels via KL-based distillation losses, guiding the student to align closely with the ensemble consensus.
\end{enumerate}

\section{Related Works}
\label{sec:related_works}

\paragraph{Self‐Supervised Learning for Visual Representation.}
Self‐supervised learning (SSL) \cite{9086055,Zbontar2021BarlowTS,Liu2020SelfSupervisedLG} has transformed visual representation learning by extracting potent features from unlabeled images. Two main frameworks underpin this progress: contrastive learning \cite{He2019MomentumCF,Chen2020ImprovedBW,Zhang2023PatchLevelCW}, which brings augmented views of the same image closer in latent space, and masked image modeling \cite{Chen2020ASF,He2021MaskedAA,Chen2022ContextAF}, which predicts occluded patches from available context. Recent extensions have pushed these paradigms further: adversarial augmentations produce harder negatives to improve robustness \cite{Kang2025Enhancing}, clustering‐based objectives enforce balanced group discrimination to avoid collapse \cite{Metaxas2024EfficientUV}, and web‐scale, vision‐only pretraining has closed the gap with language‐supervised approaches like CLIP \cite{Fan2025ScalingLV}. Multimodal strategies that combine visual and tactile signals enrich object understanding \cite{Dave2024MultimodalVR}, and embodied SSL leverages ego‐centric interactions to ground representations in both viewpoint and semantics \cite{Aubret2024SelfsupervisedVL}. More recently, co‐training vision transformers with frozen large language models has emerged as a means to infuse high‐level semantic priors into purely visual backbones \cite{kuzucu2025language}.

\paragraph{Knowledge Distillation Setup for Visual Representation.}

Knowledge distillation (KD) transfers representations from a large teacher to a compact student by minimizing divergences between their outputs or features \cite{Hinton2015DistillingTK}. Early supervised KD matched logits \cite{Hinton2015DistillingTK,Kim2020SelfKnowledgeDW,Luo_2018_ECCV} or feature maps \cite{Tang_2021_CVPR,He_2019_CVPR,liu2023self}, but these methods depend on labels and a single teacher. To adapt KD to self‐supervised and multi‐teacher scenarios, DeiT uses a distillation token to train a ViT student from a CNN teacher without extra data \cite{liu2023self}, and MoVE‐KD fuses multiple pretrained encoders via attention weights and LoRA adapters \cite{Cao2025MoVEKDKD}. Other advances include SeRKD’s superpixel‐level relational distillation \cite{Yan2025DelvingDI}, data‐free FGVC via adversarial attention \cite{Shao2023DatafreeKD}, and MaskedKD’s input masking that halves FLOPs \cite{Son2023TheRO}. Feature‐based ViTKD tailors layer‐wise losses for ViTs \cite{yang2022vitkd}, while ScaleKD demonstrates that large ViTs can teach CNN, MLP, and ViT students through cross‐attention feature mimicking \cite{Fan2024ScaleKDSV}. Hybrid designs such as HDKD share convolutional blocks between CNN teachers and Transformer students for medical imaging \cite{ElAssiouti2024HDKDHD}, and AMD automates multi‐step teacher–assistant cascades \cite{Han2024AMDAM}. Adaptive KD prevents forgetting during domain adaptation \cite{Nguyen2024AdaptiveKD}, and ISC‐DeiT adds a compact distillation head to classify interior design styles from limited data \cite{VO2024126972}.

\paragraph{Asymmetric Masking in KD Setup.}

Masked image modeling (MIM) \cite{Chen2020ASF,He2021MaskedAA,Chen2022ContextAF} demonstrates the power of random token masking for context‐based feature learning, while knowledge distillation (KD) \cite{Hinton2015DistillingTK} transfers rich representations from teacher to student. However, existing self‐supervised distillation methods \cite{Touvron2020TrainingDI,Ren2023TinyMIMAE} and supervised multi‐teacher KD approaches \cite{Anil2018LargeSD,Daquan2020RethinkingBS} typically apply the same masking ratio to both teacher and student, which constrains teachers to the student’s limited view. Asymmetric masking—in which teachers receive lighter, distinct masks than the heavily occluded student—remains largely unexplored in KD. In this work, we fill this gap by introducing a two‐way masking protocol that gives each teacher more context than the student, thereby enhancing feature interpolation and boosting the effectiveness of multi‐teacher self‐supervised distillation.

\section{Methodology}
\label{sec:method}

\subsection{Preliminaries}

\paragraph{Vision Transformers.} The Vision Transformer (ViT) \cite{Dosovitskiy2020AnII} splits an image \(x\in\mathbb{R}^{H\times W\times3}\) into \(N = \tfrac{H\,W}{P^2}\) non‐overlapping patches of size \(P=16\). For a \(224\times224\) input, this yields \(N=14\times14\) patches. Each patch \(\mathbf{p}_n\in\mathbb{R}^{16^2\cdot3}\) is projected into a \(D\)-dimensional embedding:
\[
\mathbf{z}_n^{(0)} = \mathbf{W}_{\mathrm{proj}}\,\mathbf{p}_n + \mathbf{e}_n,
\quad n=1,\dots,N,
\]
where \(\mathbf{W}_{\mathrm{proj}}\in\mathbb{R}^{D\times(16^2\cdot3)}\) and \(\mathbf{e}_n\in\mathbb{R}^D\) is a learnable positional encoding. A learnable class token \(\mathbf{z}_0^{(0)}\) is prepended, and the full sequence \(\{\mathbf{z}_n^{(\ell)}\}_{n=0}^N\) passes through \(L\) Transformer encoder blocks, each comprising multi‐head self‐attention and an MLP with residual connections and LayerNorm, yielding output tokens \(\mathbf{z}_n^{(L)}\). The final class token \(\mathbf{z}_0^{(L)}\) serves as a global representation for downstream heads.

\paragraph{ViTs for SSL distillation.} In CoMAD, the student \(E_S\) is a ViT‐Tiny model with patch size \(P=16\), embedding dimension \(D_S=192\), and depth \(L_S=12\), parameterized by \(\theta_S\). The teachers \(\{E_T^{(m)}\}\) are ViT‐Base models with the same patch size \(P=16\), embedding dimension \(D_T=768\), and depth \(L_T=12\), pretrained under MAE, MoCo v3, and iBOT , with parameters \(\{\theta_T^{(m)}\}\) held fixed. To reconcile the dimension mismatch \((D_T\to D_S)\), we employ lightweight adapters on each teacher token:
\[
\hat{\mathbf{z}}_n^{T(m)}
=\mathrm{LN}\bigl(\mathbf{W}_{\mathrm{adp}}\,\mathbf{z}_n^{T(m)}+\mathbf{b}_{\mathrm{adp}}\bigr),
\]
where \(\mathbf{W}_{\mathrm{adp}}\in\mathbb{R}^{D_S\times D_T}\), \(\mathbf{b}_{\mathrm{adp}}\in\mathbb{R}^{D_S}\), and LN denotes LayerNorm. These adapters are the only trainable components in the teacher branch, enabling the student to absorb multi‐teacher knowledge while all \(E_T^{(m)}\) remain frozen.

\subsection{Asymmetric Masking}

In CoMAD, we apply asymmetric masking to challenge the student with a heavily occluded view while granting each teacher progressively richer contexts. After patch embedding, we generate a binary mask \(M^S\) for the student that hides most tokens (keep-rate \(1 - r_S\)), and for each teacher \(m\) a lighter mask \(M^{T(m)}\) (keep-rate \(1 - r_{T(m)}\)), always preserving the class token. We then apply these masks by element-wise multiplication:
\[
\tilde Z^S = M^S \odot Z^S
\quad\text{and}\quad
\tilde Z^{T(m)} = M^{T(m)} \odot Z^{T(m)}.
\]
Since \(r_S > r_{T(m)}\), the student sees far fewer patches than any teacher, forcing it to rely on distilled signals to reconstruct missing features. By sampling each teacher’s mask independently, we further encourage diverse partial views, setting up the subsequent adapter projection and consensus gating stages.

\begin{figure*}[t]
  \centering
  \includegraphics[width=\textwidth]{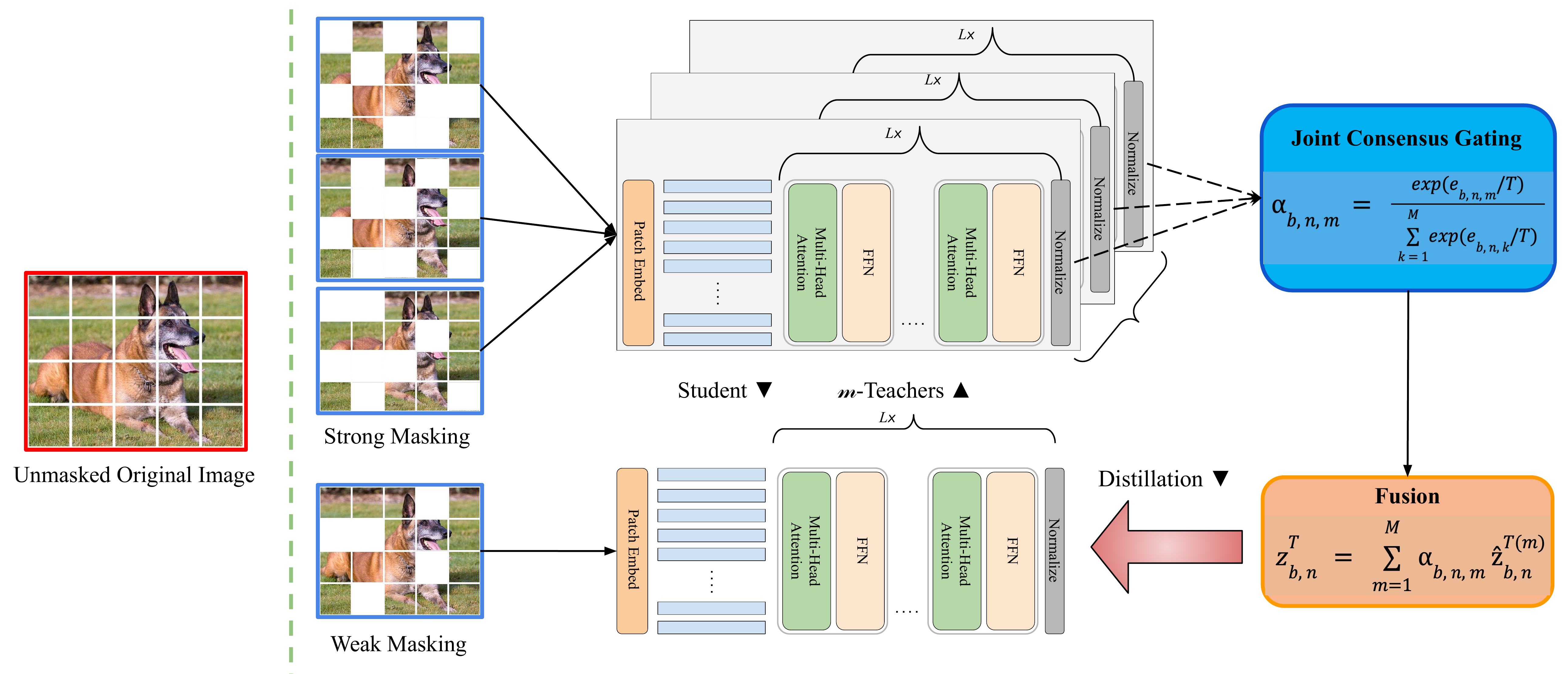}
  \caption{Illustration of the proposed CoMAD framework showing a simple series of asymmetric masking for student and multiple teachers, adapter-based projection, consensus-aware token fusion, and dual-level KL distillation.}
  \label{fig:main_arch}
\end{figure*}

\subsection{Adapter Projection}

Each teacher \(E_T^{(m)}\) produces a masked token sequence \(\mathbf{Z}^{T(m)}\in\mathbb{R}^{B\times (N+1)\times D_T}\), where \(B\) is the batch size, \(N\) the number of patches, and \(D_T=768\). To align these with the student’s tokens \(\mathbf{Z}^S\in\mathbb{R}^{B\times (N+1)\times D_S}\) (\(D_S=192\)), we apply a lightweight adapter to each teacher token:
\[
\hat{\mathbf{z}}_{b,n}^{T(m)}
=\mathrm{LayerNorm}\bigl(\mathbf{W}_{\mathrm{adp}}\;\mathbf{z}_{b,n}^{T(m)} + \mathbf{b}_{\mathrm{adp}}\bigr),
\]
where \(\mathbf{W}_{\mathrm{adp}}\in\mathbb{R}^{D_S\times D_T}\) and \(\mathbf{b}_{\mathrm{adp}}\in\mathbb{R}^{D_S}\). This precisely mirrors our \texttt{Adapter} module, consisting of a single linear projection followed by LayerNorm, and introduces negligible overhead.

Only the adapter parameters \(\{\mathbf{W}_{\mathrm{adp}},\mathbf{b}_{\mathrm{adp}}\}\) and the student weights \(\theta_S\) are updated during training; all teacher weights remain frozen. By bringing every teacher token into the same \(D_S\)-dimensional space, we ensure that the subsequent consensus gating and fusion operate over a unified embedding format, streamlining multi-teacher knowledge transfer into the compact student model.

\subsection{Joint Teachers Consensus Gating}

When distilling from multiple teachers, their predictions may conflict at a given token. To resolve this, we assign each teacher a weight that reflects both its agreement with the student’s own view and its consistency with the other teachers. Concretely, let \(z^S_{b,n}\) be the student’s embedding for sample \(b\) at position \(n\), and \(\hat z^{T(m)}_{b,n}\) the corresponding adapted embedding from teacher \(m\). We first measure their similarity by  
\[
s_{b,n,m}
\;=\;\frac{\langle z^S_{b,n},\,\hat z^{T(m)}_{b,n}\rangle}
           {\|z^S_{b,n}\|\;\|\hat z^{T(m)}_{b,n}\|},
\]  
and then gauge how well teacher \(m\) aligns with its peers via  
\[
c_{b,n,m}
\;=\;\frac{1}{M-1}\sum_{k\neq m}
    \frac{\langle \hat z^{T(m)}_{b,n},\,\hat z^{T(k)}_{b,n}\rangle}
         {\|\hat z^{T(m)}_{b,n}\|\;\|\hat z^{T(k)}_{b,n}\|}.
\]  
Summing these two terms gives a combined score  
\[
e_{b,n,m} = s_{b,n,m} + c_{b,n,m},
\]  
which we normalize across all \(M\) teachers using a softmax with temperature \(\tau\):  
\[
\alpha_{b,n,m}
= \frac{\exp\bigl(e_{b,n,m}/\tau\bigr)}
       {\sum_{k=1}^M \exp\bigl(e_{b,n,k}/\tau\bigr)}.
\]  
The resulting \(\alpha_{b,n,m}\) weights naturally down‐weight any teacher whose view diverges from the student or the ensemble, yielding a consensus‐aware fusion that is both adaptive and interpretable.

\subsection{Fusion and Distillation Losses}

Having obtained per‐token gating weights \(\alpha_{b,n,m}\) and adapted teacher embeddings \(\hat z^{T(m)}_{b,n}\), we form a single fused target for each student token by  
\[
z^T_{b,n} \;=\; \sum_{m=1}^M \alpha_{b,n,m}\,\hat z^{T(m)}_{b,n},
\]
yielding the tensor \(\mathbf{Z}^T\in\mathbb{R}^{B\times (N+1)\times D_S}\). The student’s goal is now to align its own masked embeddings \(\mathbf{Z}^S\) with this ensemble‐informed reference.

To achieve this, we introduce two complementary KL‐based losses. First, a \emph{token‐level} divergence applied only at positions the student actually observes (i.e. where its mask \(M^S_{b,n}=1\)):
\[
\mathcal{L}_{\mathrm{token}}
=\frac{1}{\sum_{b,n}M^S_{b,n}}
\sum_{b=1}^B\sum_{n=0}^N M^S_{b,n}\;\mathrm{KL}\bigl(\phi(z^S_{b,n})\;\|\;\phi(z^T_{b,n})\bigr),
\]
where \(\phi\) denotes a lightweight projection (e.g. a small MLP) that converts each \(D_S\)-dim token into a probability distribution. This term ensures the student’s visible tokens match the fused teacher consensus.

Second, a \emph{spatial‐level} divergence treats the \(N\) patch tokens (excluding the class token) as a feature map of shape \((D_S\times H'\times W')\). We compute KL divergence along the channel dimension for each spatial location and average:
\[
\mathcal{L}_{\mathrm{spatial}}
=\frac{1}{H'W'}\sum_{i,j}\mathrm{KL}\bigl(\psi\bigl(F^S_{i,j}\bigr)\,\big\|\,\psi\bigl(F^T_{i,j}\bigr)\bigr),
\]
where \(F^{S/T}\!\in\!\mathbb{R}^{B\times D_S\times H'\times W'}\) are the reshaped student and fused‐teacher features, and \(\psi\) projects each \(D_S\)-dim channel vector to a distribution.

The overall training objective combines these terms:
\[
\mathcal{L} \;=\; \mathcal{L}_{\mathrm{token}} \;+\; \mathcal{L}_{\mathrm{spatial}}.
\]
Minimizing \(\mathcal{L}\) drives the student to recover both the local token‐level semantics and the global spatial structure encoded by the ensemble of self‐supervised teachers.

\subsection{Experimental Setup}

We conduct distillation and evaluation on ImageNet-1K (1.28 M train / 50 K val), with additional tests on COCO 2017 (118 K train / 5 K val) and ADE20K (20 K train / 2 K val). All experiments run on 8 NVIDIA V100 GPUs using mixed-precision. Three ViT-Base teachers are pretrained for 300 epochs on ImageNet-1K (MAE, MoCo v3, iBOT). The CoMAD student (ViT-Tiny, 5.4 M params, embed 192, 12 layers) is loaded from a 21 K to 1 K distilled checkpoint and further distilled for 300 epochs on the unlabeled ImageNet-1K train split. We use AdamW with initial LR $1.5\times10^{-4}$, weight decay 0.05, batch size 4096, a 15-epoch linear warm-up and cosine decay. Mask ratios are 0.75 for the student and $\{0.50,\,0.40,\,0.30\}$ for the three teachers, with temperature $\tau=0.1$. Input augmentation includes random resized crop and horizontal flip for all models, plus color jitter for the student. For supervised evaluation, we fine-tune on labeled ImageNet-1K for 100 epochs (LR $1\times10^{-3}$, weight decay 0.05) under the same augmentations.

\section{Experimental Results}

\begin{table}[t!]
\centering
\small
\caption{Fine‐tune accuracy (\%) on ImageNet‐1K val set.}
\label{tab:imagenet}
\begin{tabular}{|ll|c|r|}
\toprule
Method & Teacher & Epochs & Top‐1 \\
\midrule
\multicolumn{4}{|l|}{\textbf{ViT‐T (5M)}} \\
DeiT            & Label              & 300  & 72.2 \\
MAE             & Pixel              & 1600 & 71.6 \\
MoCoV3          & EMA                & 1600 & 73.3 \\
TinyMIM         & MAE ViT‐B           & 300  & 74.6 \\
TinyMIM         & iBOT ViT‐B         & 300  & 74.6 \\
TinyMIM         & MoCoV3 ViT‐B       & 300  & 74.5 \\
DMT      & MAE/MoCoV3/iBOT    & 300  & 75.0 \\
CoMAD (ours)    & MAE/MoCoV3/iBOT           & 300    & \textbf{75.4} \\

\midrule
\multicolumn{4}{|l|}{\textbf{ViT‐S (22M)}} \\
DeiT            & Label              & 300  & 79.9 \\
MAE             & Pixel              & 1600 & 80.6 \\
MoCoV3          & EMA                & 1600 & 81.4 \\
DINO            & EMA                & 1600 & 81.5 \\
BeiT            & DALL-E             & 3200 & 81.6 \\
TinyMIM         & MAE ViT‐B           & 300  & 81.7 \\
TinyMIM         & MoCoV3 ViT‐B       & 300  & 81.5 \\
TinyMIM         & iBOT ViT‐B         & 300  & 81.5 \\
DMT      & MAE/MoCoV3/iBOT    & 300  & 82.2 \\
CoMAD (ours)    & MAE/MoCoV3/iBOT    & 300  & \textbf{82.9} \\
\midrule
\multicolumn{4}{|l|}{\textbf{ViT‐B (86M)}} \\
MAE             & Pixel              & 300  & 82.8 \\
MoCoV3          & EMA                & 800  & 83.2 \\
iBOT            & Pixel/EMA          & 300  & 83.2 \\
DMT      & MAE/MoCoV3/iBOT    & 300  & 83.7 \\
CoMAD (ours)    & MAE/MoCoV3/iBOT    & 300  & \textbf{84.7} \\
\bottomrule
\end{tabular}
\end{table}

\subsection{Main Results}

Refer to table~\ref{tab:imagenet}, where we report Top-1 accuracy on the ImageNet-1K validation set for ViT-Tiny, ViT-Small and ViT-Base students distilled under various schemes.  Apart from single-teacher baselines (TinyMIM-MAE, TinyMIM-MoCo v3, TinyMIM-iBOT), we compare with the original DMT and our CoMAD.  For ViT-Tiny, CoMAD improves Top-1 accuracy from 75.0 \% to 75.4 \%.  Similar gains are observed for ViT-Small (+0.3 pp) and ViT-Base (+0.5 pp), demonstrating consistent benefits of consensus gating and asymmetric masking.

\subsection{Transfer Learning Tests}

To evaluate CoMAD’s transferability, we plug the distilled ViT-Tiny backbone into a Cascade R-CNN with FPN on MS-COCO and a UPerNet head on ADE20K, and summarize results in Table~\ref{tab:transfer}. For COCO we report box AP (\(\mathrm{AP}^{bb}\)) and mask AP (\(\mathrm{AP}^{mk}\)) at IoU thresholds 0.5:0.95. For ADE20K we report mean IoU (mIoU) and pixel accuracy (aAcc). Compared to the original DMT student, CoMAD boosts ADE20K mIoU from 46.9 \% to 47.3 \% and aAcc from 82.9 \% to 83.1 \%, and raises COCO AP\(^ {bb}\) from 44.3 \% to 44.5 \% and AP\(^ {mk}\) from 40.3 \% to 40.5 \%, confirming stronger dense-prediction performance under our consensus-gated, asymmetrically masked distillation.

\begin{table}[t!]
\centering
\small
\caption{Results on ADE20K and MS-COCO with ViT‑S. TinyMIM is abbreviated as “TM.”}
\label{tab:transfer}
\resizebox{\columnwidth}{!}{%
\begin{tabular}{|l|cc|cc|}
\toprule
Method            & \multicolumn{2}{c|}{ADE20K} & \multicolumn{2}{c|}{MS-COCO} \\
                  & mIoU        & aAcc        & AP\textsuperscript{bb}       & AP\textsuperscript{mk}       \\
\midrule
DeiT \cite{Touvron2020TrainingDI}      & 43.1        & –           & 43.1                         & 38.4                         \\
MAE \cite{He2021MaskedAA}           & 42.8        & –           & –                            & –                            \\
MoCoV3 \cite{Chen2021AnES}       & 43.9        & –           & 39.8                         & 37.1                         \\
iBOT \cite{Zhou2021iBOTIB}         & 44.1        & 81.4        & 42.6                         & 39.0                         \\
DINO \cite{Caron_2021_ICCV}         & 42.3        & 80.4        & 40.8                         & 37.3                         \\
ADCLR \cite{Zhang2023PatchLevel}         & 44.2        & 81.8        & 43.8                         & 39.2                         \\
PQCL \cite{10.5555/3618408.3620173}         & 45.2        & 81.9        & 43.1                         & 39.3                         \\
TM-MAE \cite{Ren2023TinyMIMAE}       & 44.8        & 81.9        & 41.4                         & 38.1                         \\
TM-MoCoV3 \cite{Ren2023TinyMIMAE}    & 45.3        & 82.0        & 42.1                         & 38.9                         \\
TM-iBOT \cite{Ren2023TinyMIMAE}      & 46.0        & 82.3        & 43.1                         & 39.6                         \\
DMT \cite{Liu2023DMTCD}       & 46.9        & 82.9        & 44.3                         & 40.3                         \\
\textbf{CoMAD (Ours)}      & \textbf{47.3} & \textbf{83.1} & \textbf{44.5}              & \textbf{40.5}                \\
\bottomrule
\end{tabular}%
}
\end{table}

\subsection{Ablation Study}

\paragraph{Effect of Distillation Losses.}  
We compare variants using only token‐fusion distillation (TFD) or spatial‐fusion distillation (SFD), their combination, and an averaged “m-TinyMIM” mean‐fusion baseline (Table~\ref{tab:ablate-loss}). Single‐loss models offer modest mIoU/aAcc gains over the TinyMIM baseline, but combining TFD and SFD captures both local token semantics and global feature distributions, yielding a substantial boost. Replacing KL divergence with MSE also hurts accuracy, underscoring the importance of matching full distributions. CoMAD adopts the dual‐KL design, which achieves the highest segmentation performance.

\begin{table}[t]
\centering
\small
\caption{ADE20K mIoU/aAcc and COCO APs with different distillation losses.}
\label{tab:ablate-loss}
\begin{tabular}{|l|cc|cc|}
\toprule
Loss terms           & KL & MSE & mIoU & aAcc  \\
\midrule
TinyMIM (baseline)   &    &     & 39.2  & 79.9  \\
\;+\;TFD only        & \checkmark &     & 34.5  & 78.2  \\
\;+\;SFD only        &     & \checkmark & 41.6  & 80.7  \\
m-TinyMIM (mean)     & \checkmark &     & 41.7  & 80.7  \\
Dual w/ MSE          &     & \checkmark & 41.7  & 80.8  \\
Dual w/ KL (ours)    & \checkmark &     & \textbf{42.0} & \textbf{81.2} \\
\bottomrule
\end{tabular}
\end{table}

\paragraph{Combination of Teachers.}  
We next isolate the impact of each teacher (MAE, MoCo v3, iBOT) by distilling from one, two, or all three under our dual‐KL loss (Table~\ref{tab:ablate-teachers}). Any pair already outperforms single‐teacher distillation, (e.g.\ MAE+MoCo v3 adds +1.6 pp mIoU) while including all three yields the best result. This confirms that our consensus‐aware fusion effectively reconciles diverse self‐supervised signals into a stronger student.

\begin{table}[h]
\centering
\small
\caption{ADE20K mIoU/aAcc for different teacher sets.}
\label{tab:ablate-teachers}
\begin{tabular}{|ccc|cc|}
\toprule
MAE & MoCoV3 & iBOT & mIoU & aAcc \\
\midrule
\checkmark &         &        & 39.2 & 79.9 \\
           & \checkmark &      & 40.3 (+1.1) & 80.5 \\
           &          & \checkmark & 40.0 (+0.8) & 80.4 \\
\checkmark & \checkmark &      & 40.8 (+1.6) & 80.4 \\
\checkmark &          & \checkmark & 40.2 (+1.0) & 80.1 \\
\checkmark & \checkmark & \checkmark & \textbf{42.0 (+2.8)} & \textbf{81.2} \\
\bottomrule
\end{tabular}
\end{table}

\paragraph{Mask Ratio Settings.}  
Choosing mask ratios balances challenge and guidance: too much masking starves the student, too little reduces the distillation task. Table~\ref{tab:ablate-mask} sweeps the student keep‐rate (\(1-r_S\)) from 20 \% to 30 \% alongside two teacher schedules. The default setting (\(r_S=0.75\), \(r_T=\{0.50,0.40,0.30\}\)) delivers peak Top‐1 and mIoU, while deviations in either direction degrade performance, validating our asymmetric masking design.

\begin{table}[h]
\centering
\small
\caption{Mask‐ratio ablation (ViT-Tiny).}
\label{tab:ablate-mask}
\begin{tabular}{|c|ccc|cc|}
\toprule
$r_S$ & $r_T^{(1)}$ & $r_T^{(2)}$ & $r_T^{(3)}$ & IN-1K Top-1 & ADE20K mIoU \\
\midrule
0.80 & 0.45 & 0.35 & 0.25 & 74.7  & 41.5 \\ 
0.75 & 0.50 & 0.40 & 0.30 & \textbf{75.3} & \textbf{42.0} \\ 
0.70 & 0.55 & 0.45 & 0.35 & 74.5  & 41.7 \\ 
\bottomrule
\end{tabular}
\end{table}

\paragraph{Gating Mechanism.}  
Finally, we dissect our joint consensus gating into its affinity and inter‐teacher components (Table~\ref{tab:ablate-gate}). A uniform average already improves over naïve mean fusion, but using only affinity or only consensus yields smaller gains. Only the full gating, combining both student–teacher alignment and teacher–teacher agreement, achieves the largest Top‐1 and mIoU improvements, demonstrating the necessity of both signals for robust multi‐teacher fusion.

\begin{table}[h]
\centering\small
\caption{Ablation of the gating mechanism on ViT‐Tiny. ``Aff.'' denotes student–teacher affinity and ``Cons.'' denotes inter‐teacher consensus. Performance is measured by Top‐1 accuracy on ImageNet‐1K and mIoU on ADE20K.}
\label{tab:ablate-gate}
\begin{tabular}{@{}l|cc|cc@{}}
\toprule
Variant                     & \multicolumn{2}{c|}{Gating Terms}   & \multicolumn{2}{c}{Performance} \\
                            & Aff.      & Con.         & IN‐1K (\%)    & ADE20K (\%)   \\
\midrule
Uniform (no gating)         &           &              & 74.5          & 41.3          \\
Aff. only                & \checkmark&              & 74.8          & 41.7          \\
Cons. only               &           & \checkmark   & 74.6          & 41.5          \\
Combined (ours)              & \checkmark& \checkmark   & \textbf{75.4} & \textbf{42.0} \\
\bottomrule
\end{tabular}
\end{table}

\section{Conclusion}

In this paper, we have introduced Consensus‐oriented Masked Distillation (CoMAD), a lightweight framework that unifies the strengths of multiple self‐supervised ViT‐Base teachers into a compact student model. By combining asymmetric masking, in which the student observes only a fraction of patches while each teacher receives progressively lighter and distinct masks, with a non‐parametric joint consensus gating mechanism and dual‐level KL distillation on both token embeddings and spatial feature maps, CoMAD effectively captures complementary semantic and contextual cues. Our extensive evaluation on ImageNet‐1K classification as well as ADE20K and MS‐COCO dense prediction benchmarks shows that CoMAD consistently outperforms single‐teacher state-of-the-art baselines without introducing any extra learnable parameters. These results demonstrate that multi‐teacher fusion in self‐supervised learning can yield richer student representations and practical efficiency for deployment. Future works can explore adaptive mask schedules and the inclusion of heterogeneous teacher architectures to further enrich distilled representations.

\bibliography{AAAI_26}

\end{document}